\PassOptionsToPackage{table}{xcolor}
\documentclass[letterpaper, 10 pt, conference]{ieeeconf}  
\usepackage[dvipsnames,table,xcdraw]{xcolor}
\usepackage{cite}
\usepackage{amsmath,amssymb,amsfonts}
\usepackage[graphicx]{realboxes}
\usepackage{textcomp}
\usepackage{multirow}
\usepackage{array}
\usepackage{tikz}
\usepackage{siunitx}
\usepackage{lscape}
\usepackage{adjustbox}
\usepackage{tabularx}
\usepackage{mleftright}
\usepackage{etoolbox}
\usepackage{amsmath}
\usepackage{amssymb}
\usepackage{tablefootnote}
\usepackage{verbatim}
\usepackage{graphicx}
\usepackage{booktabs}
\usepackage{mathtools}
\usepackage{makecell}

\usepackage{algorithm}
\usepackage{algpseudocode}

\IEEEoverridecommandlockouts                              

\title{\LARGE \bf
RaDL: Relation-aware Disentangled Learning for Multi-Instance Text-to-Image Generation
}

\author{Geon Park$^{1}$, Seon Bin Kim$^{1}$, Gunho Jung$^{1}$, and Seong-Whan Lee$^{1}$
\thanks{*This research was supported by the Institute of Information \& Communications Technology 
Planning \& Evaluation (IITP) grant, funded by the Korea government (MSIT) (No. RS-2019-
II190079 (Artificial Intelligence Graduate School Program (Korea University)), and No. IITP-2025-
RS-2024-00436857 (Information Technology Research Center (ITRC)).}
\thanks{$^{1}$G. Park, S. B. Kim, G. Jung, and S.-W. Lee are with the Department of Artificial Intelligence, Korea University, Anam-dong, Seongbuk-ku, Seoul 02841, Korea.
{\tt\small \{geonpark, s\_b\_kim, gh\_jung, sw.lee\}@korea.ac.kr}
}
}

\begin{document}

\maketitle
\thispagestyle{empty}
\pagestyle{empty}


\begin{abstract}

With recent advancements in text-to-image (T2I) models, effectively generating multiple instances within a single image prompt has become a crucial challenge. Existing methods, while successful in generating positions of individual instances, often struggle to account for relationship discrepancy and multiple attributes leakage. To address these limitations, this paper proposes the relation-aware disentangled learning (RaDL) framework. RaDL enhances instance-specific attributes through learnable parameters and generates relation-aware image features via Relation Attention, utilizing action verbs extracted from the global prompt. Through extensive evaluations on benchmarks such as COCO-Position, COCO-MIG, and DrawBench, we demonstrate that RaDL outperforms existing methods, showing significant improvements in positional accuracy, multiple attributes consideration, and the relationships between instances. Our results present RaDL as the solution for generating images that consider both the relationships and multiple attributes of each instance within the multi-instance image.

\end{abstract}

\section{INTRODUCTION}

In recent years, with the advancement of text-to-image (T2I) generation models, research on the multi-instance generation (MIG) task, which generates multiple instances within a single image, has been actively performed. This enables the construction of complex multi-instance images from a single prompt, leading to more expressive visual representations. However, when generating images containing multiple instances, existing methods \cite{pmlr-v139-ramesh21a, rombach2022high, saharia2022photorealistic} have limitations in generating unique attributes and precise locations of each instance using a single text prompt.

To overcome these limitations, recent research explores layout-to-image (L2I) approaches \cite{yang2023reco, lim2000text, zhang2023text2layer} that explicitly provide location information for individual instances. Unlike existing T2I models \cite{Zhao_2023_ICCV, lee2018deep, qiao2019mirrorgan} that rely solely on textual descriptions, L2I methods utilize structured layouts such as bounding boxes and segmentation masks. Specifically, some frameworks \cite{liu2022compositional, zhou2024migc} that are based on the divide and conquer (DAC) approach disentangle multiple instances into individual single ones for training each instance independently before integrating them into a unified whole.

\begin{figure}
    \centering
    \includegraphics[width=\linewidth]{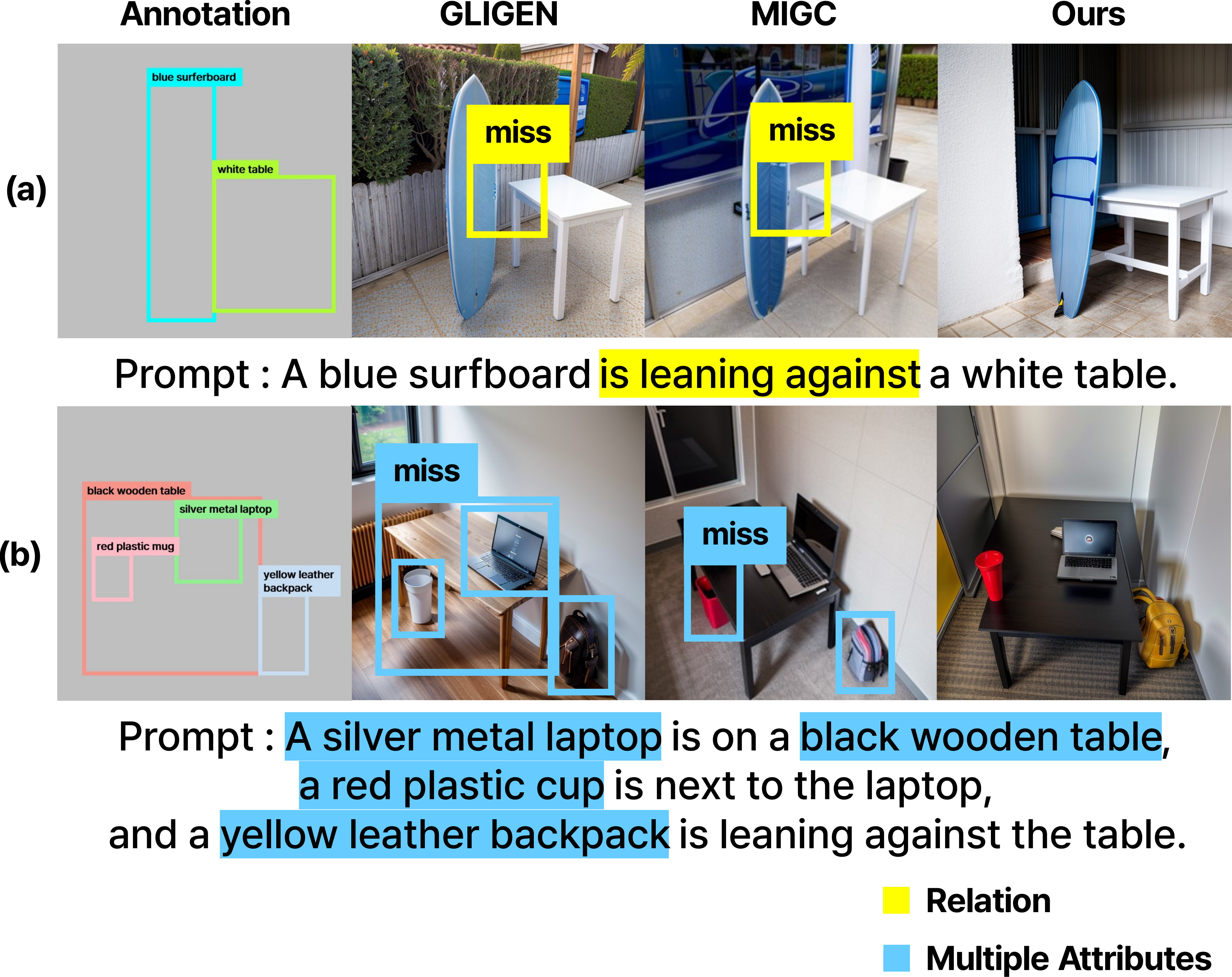}
    \caption{An illustration of L2I model outputs reveals limitations in representing the relationships of each instance and preserving multiple attributes. (a) Relationship discrepancy occurs when the model fails to consider spatial or interactive relationships between instances. (b) Multiple attributes leakage occurs when intended attributes, such as color or material, are missed or only partially applied.}
    \label{fig:1}
\end{figure}

Although DAC-based methods alleviate the issue of missing instances in generated images, they remain limited by relationship discrepancy and multiple attributes leakage. Relationship discrepancy refers to the problem where semantic relationships between instances are not represented during image generation. This issue arises from the DAC-based processing, where instances are separated and trained independently, resulting in the omission of relative information such as mutual position, direction, and actions, which are essential for constructing relationships. Multiple attributes leakage refers to the problem where the multiple attributes information described in the text is lost, resulting in unintended attributes being applied to the image or only a subset of the intended attributes being applied. This leads to entangled representations, as multiple attributes described in a single sentence are processed through a shared attention space, making it difficult to assign each attribute to its corresponding visual region. To illustrate this, Fig. \ref{fig:1} shows the limitations in representing the relationships of each instance and preserving multiple attributes. In Fig. \ref{fig:1}(a), existing L2I models consider attributes and generate images with a \textit{blue surfboard} and a \textit{white table} but do not consider the relationships \textit{leaning against} between instances, as shown in the yellow boxes. In Fig. \ref{fig:1}(b), when there are numerous layouts and attributes, recent models tend to consider only some attributes or miss entirely, leading to multiple attributes leakage, as seen in the instances within the blue boxes.

 To address relationship discrepancy and multiple attributes leakage in multi-instance image generation, we propose the relation-aware disentangled learning (RaDL) framework. We follow the DAC-based approach to perform multi-instance disentanglement and semantic instance fusion. Specifically, multi-instance disentanglement separates each instance using its mask and enhances image features to prevent attribute degradation. The image features of each instance are considered in a form that emphasizes attribute information through learnable parameters. By applying cross-attention between text label and emphasized image features, the model considers multiple attributes and preserves unique visual information during training.

In semantic instance fusion, we utilize verbs that express interactions or spatial relations between instances to understand their relationships \cite{hoe2024interactdiffusion}. Verb information extracted from the global prompt is combined with image features through cross-attention with a total mask applied via element-wise multiplication to express relationships effectively. Background and all instance features are combined using a normalization method (\emph{i.e.}, softmax) that adjusts importance through mask-based weights to generate the entire feature map. This enables image generation that applies the unique attributes of each instance during fusion while considering relationships between them. In Fig. \ref{fig:1}, RaDL generates multi-instance images, considering both relationships and multiple attributes between instances. In the first example (a), it considers the relationships between instances where the surfboard leans against the table. In the second example (b), it generates instances that apply all intended attributes, even in cases with multiple attributes like \textit{silver metal laptop}, during the image generation process.

Our contributions are summarized as follows:
\begin{itemize}
 \item We propose RaDL that emphasizes multiple attributes of each instance in multi-instance image generation. Using learnable parameters, multiple attributes for each instance are considered, and unique visual attributes can be maintained during training.

 \item We introduce Relation Attention that considers relationships between instances. It emphasizes interactive relationships by utilizing action verbs from the global prompt, enabling dynamics-driven, relationship-aware image generation.

 \item We demonstrate the effectiveness of RaDL through its superior performance compared to existing methods in various benchmarks, particularly focusing on instance relationships and retaining attributes of multiple instances.
 
\end{itemize}

\section{RELATED WORKS}

\subsection{Text-to-Image Models}

T2I models refer to technologies that generate realistic images based on textual descriptions. Recently, research has been actively conducted based on diffusion models, with representative methods including Imagen \cite{saharia2022photorealistic} and Stable Diffusion \cite{rombach2022high}. Diffusion models operate by generating images through a gradual noise removal process \cite{cho2021neurograsp}. In particular, the latent diffusion model (LDM) performs forward and reverse processes in a latent space based on the variational autoencoder (VAE)  \cite{VAE}, increasing computational efficiency by addressing lower-dimensional data. Additionally, models that incorporate CLIP \cite{pmlr-v139-radford21a} enhance the understanding of textual descriptions and improve alignment between text and image representations, enabling more controllable image generation. However, existing models have limitations in expressing positional information of instances to be generated because they only use simple textual prompt inputs. Therefore, we utilize L2I to resolve this limitation of existing T2I models that cannot generate instances at desired locations using only prompts.

\begin{figure*}[!tb]
    \centering
    \includegraphics[width=0.9\textwidth]{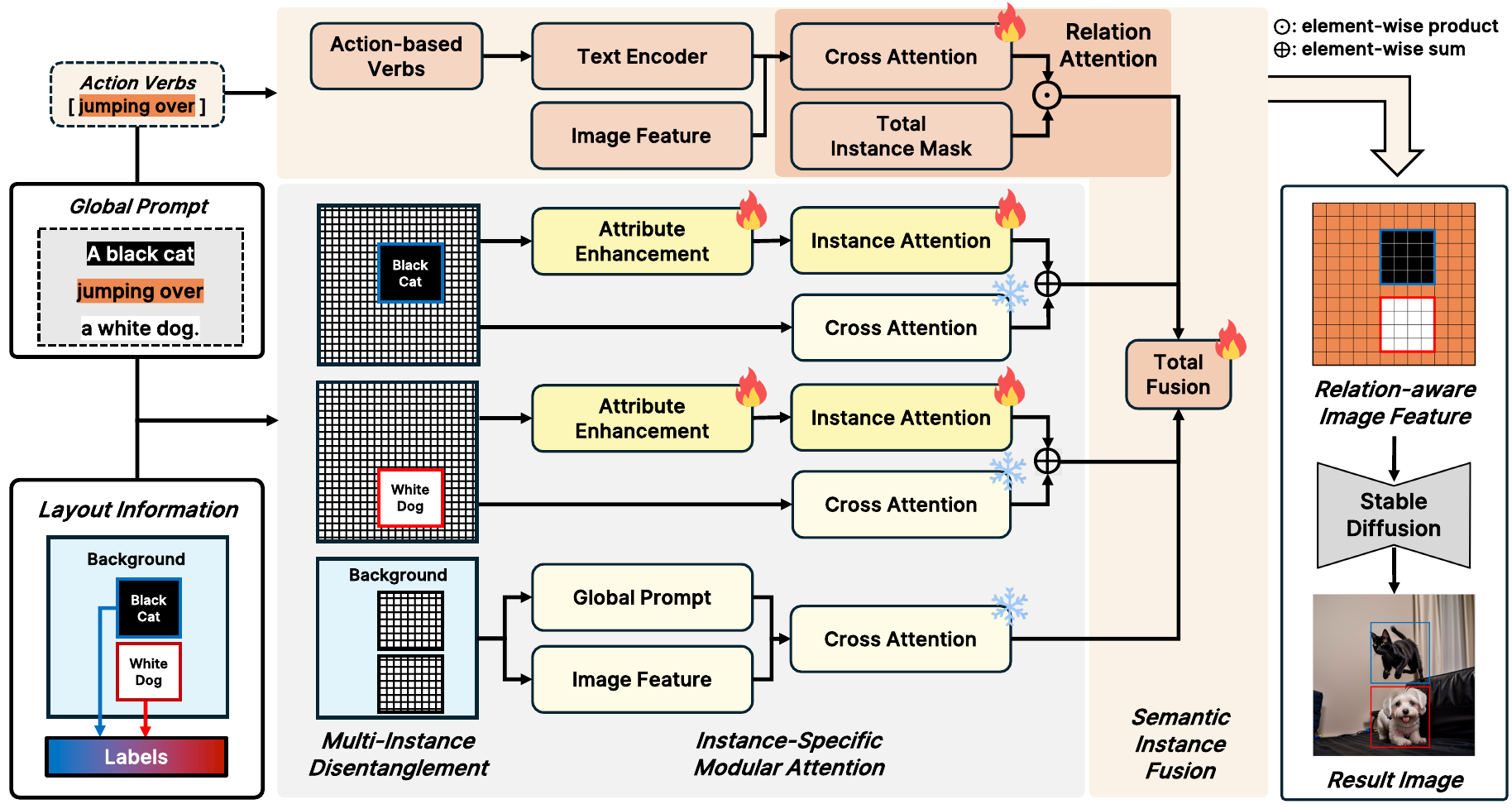}
    \caption{ Overall architecture of our proposed RaDL. RaDL enables image generation that considers both multiple attributes of individual instances and the relationships between instances. RaDL comprises three steps: (a) Multi-Instance Disentanglement (b) Instance-Specific Modular Attention (c) Multi-Stage Semantic Instance Fusion.}
    \label{fig:2}
\end{figure*}

\subsection{Image Generation from Layouts}

Image generation from layouts is a technology that generates images based on predefined instance placement information, developed to overcome spatial representation problems in existing T2I models. Generally, it provides spatial information such as bounding boxes, masks, and keypoints as conditions, which leads to generating images that more accurately consider user intentions based on this additional information \cite{yang2007reconstruction, lee1999integrated}. GLIGEN \cite{li2023gligen} is a method that directly takes bounding boxes as input to place specific instances at corresponding locations, while BoxDiff \cite{xie2023boxdiff} proposes an approach that enables more precise instance placement by combining bounding box and text conditions. Additionally, MIGC \cite{zhou2024migc} proposes a method that enables multi-instance image generation at accurate positions without instance omission by utilizing the DAC method to train multiple instances individually. All three methods improve the limitations of spatial representation in existing text-based image generation models. However, L2I models have limitations in not considering relationships and multiple attributes between multiple instances while independently defining and placing the position and shape of instances. To address the limitations of existing L2I models, we propose the Relation Attention and Attribute Enhancement Module.

\section{METHOD}

In this section, we introduce RaDL for effective multi-instance image generation that considers relationships and multiple attributes for each instance during training. The overall process of RaDL consists of three stages: multi-instance disentanglement, Instance-Specific Modular Attention, and multi-stage semantic instance fusion. The overall framework is illustrated in Fig. \ref{fig:2}.

\subsection{Problem Definition}
Multi-instance image generation produces images based on three user-defined inputs: (1) a global prompt $\mathbf{P}$ describing the overall context of the image, (2) a set of instance layout bounding boxes $\mathbf{B} = \{b^1, ..., b^N\}$, where each ${b^i} = [x_{1}^{i}, y_{1}^{i}, x_{2}^{i}, y_{2}^{i}]$ defines the position of instance $i$, and (3) instance labels $\mathbf{L} = \{l^1, ..., l^N\}$ describing the visual attributes or content of each instance.

\subsection{Multi-Instance Disentanglement}

To effectively train the attributes of each instance by disentangling them individually, we convert the labels $\mathbf{L}$ describing each instance into embeddings through the CLIP text encoder and apply cross-attention with an instance mask between the embedded labels and the instance image features. The cross-attention $\mathbf{R}^i$ is defined by the following equation:

\begin{equation}
\mathbf{R}^i = \text{softmax}\left( \frac{\mathbf{Q} \mathbf{K}_{text}^{i^\top}}{\sqrt{d}} \right)\mathbf{V}_{text}^i \odot \mathbf{M}^i,
\end{equation}

\noindent where key $\mathbf{K}_{text}^i$ and value $\mathbf{V}_{text}^i$ are obtained from the embedded labels, and the query $\mathbf{Q}$ is derived from the image feature map. $\mathbf{M}^i$ is a mask generated based on the bounding box $b^i$ of the instance, where values inside the instance region are 1. The mask constrains training to occur only within the designated area of the instance. Through this approach, we can train multiple instances in parallel by disentangling them into single instances.

\begin{table*}[!ht]
    \begin{center}
    \caption{Quantitative performance on the COCO-Position dataset}
        \label{tab:1}%
        \resizebox{\textwidth}{!}{\begin{tabular}{c| c c c c c | c c | c }
        \toprule
         \multirow{2}{*}{\textbf{Method}} & \multicolumn{5}{c|}{\textbf{Spatial Accuracy(\%)}}  & \multicolumn{2}{c|}{\textbf{Image Text Consistency}}  & \multicolumn{1}{c}{\textbf{Image Quality}}\\

    \cmidrule(lr){2-6} \cmidrule(lr){7-8} \cmidrule(lr){9-9} 
    
    & R $\uparrow$ & mIoU $\uparrow$ & AP $\uparrow$ & AP50 $\uparrow$ & AP75 $\uparrow$ & CLIP $\uparrow$ & Local CLIP $\uparrow$ & FID-6K $\downarrow$  \\
    \midrule
    Real Image & 83.75 & 85.49 & 65.97 & 79.11 & 71.22 & 24.22 & 19.74 & -  \\
    \midrule
    Stable Diffusion \cite{rombach2022high} & 5.95 & 21.60 & 0.80 & 2.71 & 0.42 & \textbf{25.69} & 17.34 & 23.56  \\
    GLIGEN \cite{li2023gligen} & 70.52 & 71.61 & 40.68 & 68.26 & 42.85 & 24.61 & 19.69 & 26.80   \\
    LayoutDiffusion \cite{zheng2023layoutdiffusion} & 50.53 & 57.49 & 23.45 & 48.10 & 20.70 & 18.28 & 19.08 & 25.94  \\
    BoxDiff \cite{xie2023boxdiff} & 17.84 & 33.38 & 3.29 & 12.27 & 1.08 & 23.79 & 18.70 & 25.15  \\
    TFLCG \cite{chen2024training} & 13.54 & 28.01 & 1.75 & 6.77 & 0.56 & 25.07 & 17.97 & 24.65  \\
    MultiDiffusion \cite{pmlr-v202-bar-tal23a} & 23.86 & 38.82 & 6.72 & 18.65 & 3.63 & 22.10 & 19.13 & 33.20    \\
    MIGC \cite{zhou2024migc} & 80.29 & 77.38 & 54.69 & 84.17 & 61.71 & 24.66 & 20.25 & 24.52 \\
    \midrule
    \rowcolor{Gray!25}
    Ours & \textbf{81.09} & \textbf{78.58} & \textbf{55.47} & \textbf{85.02} & \textbf{62.33} & 24.66  & \textbf{20.27} &  \textbf{23.53} \\
    \bottomrule
        \end{tabular}}
    \end{center}
\end{table*}

\subsection{Instance-Specific Modular Attention}

After disentangling multiple instances individually, we employ Attribute Enhancement to maintain and strengthen multiple attributes of each instance. The correspondence between specific words in the label and image regions is trained as learnable parameters in the framework, producing an image feature map with emphasized attributes for each instance. This allows for effective training while maintaining multiple attributes of each instance. The Attribute Enhancement $\mathbf{R}_{AE}^i$ is defined by the following equation:

\begin{equation}
\mathbf{R}_{AE}^i = \text{softmax}\left( \frac{\mathbf{Q}_{lp} \mathbf{K}_{img}^{i^{\top}}}{\sqrt{d}} \right)\mathbf{V}_{img}^i,
\end{equation}

\noindent where key $\mathbf{K}_{img}^i$ and value $\mathbf{V}_{img}^i$ are obtained from the image features of each instance, and the query $\mathbf{Q}_{lp}$ is a learnable parameter. To clarify the spatial location of each instance, we utilize bounding boxes $B$ as input to a multi-layer perceptron (MLP), which transforms them into a position embedding. This position embedding is then concatenated with the corresponding textual label embedding, resulting in a $\mathbf{E}^i$. This ensures that instances of the same class can be distinguished when they exist in different positions.

The enhanced image feature map is combined with the embedded representation through cross-attention with the instance mask $\mathbf{M}^i$. The Instance Attention $\mathbf{R}_{IA}^i$ is defined by the following equation:

\begin{equation}
\mathbf{R}_{IA}^i = \text{softmax}\left( \frac{\mathbf{Q}_{AE} \mathbf{K}_{E}^{i^{\top}}}{\sqrt{d}} \right)\mathbf{V}_{E}^i \odot \mathbf{M}^i,
\end{equation}

\noindent where key $\mathbf{K}_{E}^i$ and value $\mathbf{V}_{E}^i$ are obtained from $\mathbf{E}^i$, and the query $\mathbf{Q}_{AE}$ is derived from the image feature map $\mathbf{R}^i_{AE}$, which emphasizes the multiple attributes information of each instance. Through this process, Instance Attention forms a new cross-attention layer $\mathbf{R}_f^i$ by adding to the pre-trained cross-attention output $\mathbf{R}^i$, as shown in the following equation:

\begin{equation}
\mathbf{R}_f^i = \mathbf{R}^i + \mathbf{R}_{IA}^i.
\end{equation}

This preserves pre-trained semantic information while enabling training based on accurate position and multiple attributes of each instance, thereby preventing attribute degradation in instances.

For background regions, we consider the context of the entire image to generate the overall semantic flow by utilizing the global prompt $\mathbf{P}$. We integrate masks of all instances to define background areas not included in any instance. The equation for integrating all instances is as follows:

\begin{equation}
\mathbf{M}_{\text{total}}(x, y) = 
\begin{cases} 
1, & \text{if } \sum_{i=1}^{N} \mathbf{M}^i(x, y) > 0 \\
0, & \text{otherwise}
\end{cases}
.
\end{equation}

This equation determines a binary mask indicating whether each pixel $(x, y)$ in the image is a part of any instance. If the sum of all instance masks $\mathbf{M}^i$ at a given coordinate is bigger than 0, the pixel is assigned a value of 1. Otherwise, it is set to 0, representing the background.

\subsection{Multi-Stage Semantic Instance Fusion}

We perform the process of fusing image features of individually trained instances while considering their relationships. Relation Attention effectively considers relationships between instances in the image by utilizing the global prompt $\mathbf{P}$. It consists of verb sequences extracted from the global prompt that center on relationships, and it applies the CLIP text encoder to convert them into text embeddings. This allows the consideration of the overall context and relationships between instances by applying cross-attention with a total mask between the converted embeddings and the entire image features. The Relation Attention $\mathbf{R}_\text{relation}$ is expressed by the following equation:

\begin{equation}
\mathbf{R}_\text{relation} = \text{softmax}\left( \frac{\mathbf{Q} \mathbf{K}_r^{\top}}{\sqrt{d}} \right)\mathbf{V}_r \odot \mathbf{M}_\text{total},
\end{equation}

\begin{table}[!h]
\centering
\caption{Quantitative results in COCO-MIG dataset}
\begin{tabularx}{\linewidth}{c|*{2}{>{\centering\arraybackslash}X}|*{2}{>{\centering\arraybackslash}X}}
\toprule
\multirow{2}{*}{\textbf{Method}} & \multicolumn{2}{c|}{\textbf{Success Rate(\%)} $\uparrow$} & \multicolumn{2}{c}{\textbf{mIoU} $\uparrow$} \\
\cmidrule(lr){2-3} \cmidrule(lr){2-3} \cmidrule(lr){4-5}
  & ${L_2}$ & ${L_6}$ & ${L_2}$ & ${L_6}$ \\
\midrule
Stable Diffusion & 6.87 & 2.21 & 18.92 & 14.42 \\
GLIGEN & 42.30 & 30.84 & 37.58 & 27.70 \\
BoxDiff & 24.61 & 9.31 & 32.64 & 21.19 \\
TFLCG & 20.47 & 4.36 & 29.34 & 17.86 \\
MultiDiffusion & 24.88 & 18.60 & 29.41 & 24.71 \\
MIGC & 67.70 & 56.88 & 59.39 & 49.89 \\
\midrule
\rowcolor{Gray!25}
Ours & $\textbf{68.97}$ & $\textbf{59.45}$ & $\textbf{61.16}$ & $\textbf{53.27}$ \\
\bottomrule
\end{tabularx}
\label{tab:2}
\end{table}

\noindent where key $\mathbf{K}_r$ and value $\mathbf{V}_r$ are obtained from verb sequence embeddings, and the query $\mathbf{Q}$ is derived from the image feature map. To consider only valid regions where instances actually exist, the total instance mask $\mathbf{M}_{\text{total}}$ is utilized. Through this process, relation features that consider the relationships between instances within the image are generated. 

In total fusion, we normalize attention weights across the background and instances via pixel-wise softmax. This normalization process generates relation-aware image features that consider instance relationships and multiple attributes, enabling the efficient arrangement of background and instances.

\begin{figure*}
    \centering
    \includegraphics[width=0.9\linewidth]{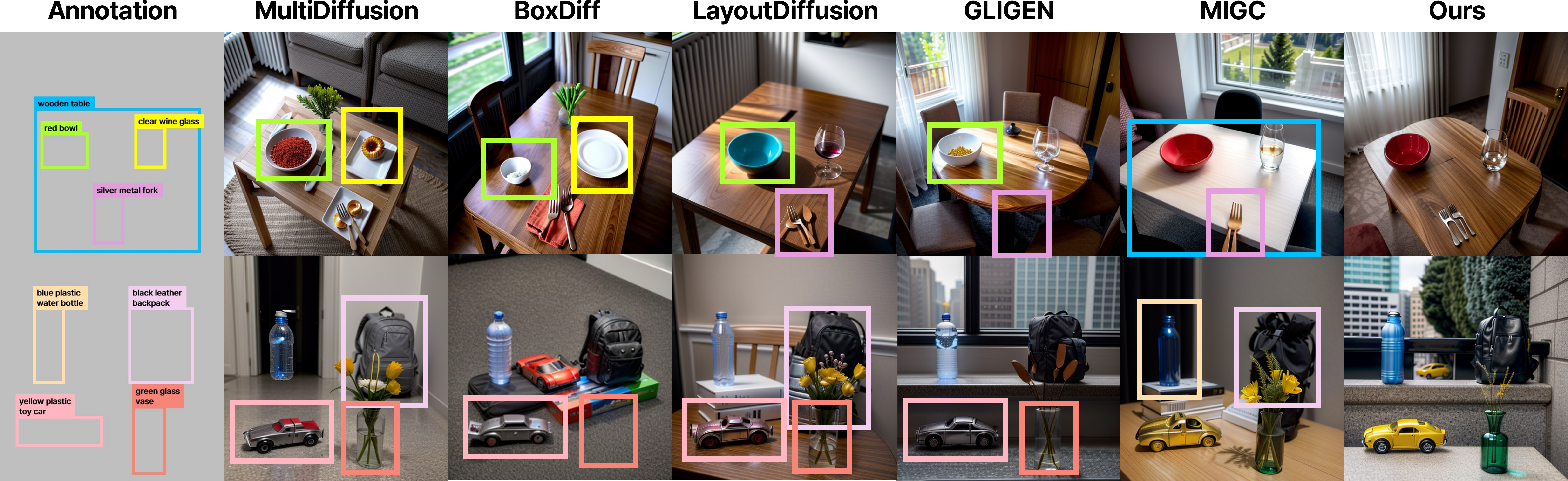}
    \caption{Qualitative comparison of RaDL and other baselines on COCO-Position dataset. The color boxes represent instances that fail to consider multiple attributes. Experimental results show that our method accurately considers multiple attributes, such as color and material corresponding to the labels, while ensuring the precise placement of instances.}
    \label{fig:3}
\end{figure*}

\section{Experiments}
\subsection{Experimental Settings}
\subsubsection{Evaluation Dataset}
We use COCO-Position \cite{lin2014microsoft}, COCO-MIG \cite{zhou2024migc}, and DrawBench datasets \cite{saharia2022photorealistic}. 

\textbf{COCO-Position dataset} is constructed based on the COCO 2014 dataset, which randomly selects 800 images. We generate 6400 images using the image caption as a global prompt, instance labels as instance descriptions, and bounding boxes as layouts.

\textbf{COCO-MIG dataset} is a multi-instance generation dataset based on COCO-Position. Each object has explicitly specified color attributes, and the dataset is categorized according to the number of instances to be generated.

\textbf{DrawBench dataset} consists of 60 prompts for text-based image generation across three conditions: color, quantity, and relationships between objects. Among these, 20 are color-related, 20 are quantity-related, and 20 are related to relationships between objects.

\subsubsection{Implementation Details}
We apply RaDL to the bottleneck (8 × 8) and low-resolution upsampling path (16 × 16) of UNet \cite{ronneberger2015u}. Model training is conducted based on pre-trained Stable Diffusion v1.4. For optimization, we use the AdamW optimizer \cite{loshchilov2018decoupled} with a learning rate that starts at 0 with linear warmup and is fixed at $1e^{-4}$. For inference, we generate images over 60 steps. During the initial 30 steps, RaDL is activated to inject relation-aware image features into the U-Net based on user-provided instance layouts and descriptions, enabling the model to consider the relationships of each instance and multiple attributes. In the latter 30 steps, only regular denoising processes are performed to refine the image details.

\subsubsection{Evaluation Metrics}

On the COCO-Position dataset, we assess spatial accuracy using the Success Rate ($\text{R}$), where all instances in an image must have an intersection over union (IoU) of 0.5 or higher and match their specified color attributes to be considered successful. We also assess mIoU and detection metrics (AP, AP50, and AP75) using GroundingDINO \cite{liu2024grounding}. Text-image alignment is measured using the CLIP score and Local CLIP score, while overall image quality is evaluated using FID \cite{heusel2017gans}. On the COCO-MIG dataset, we adopt the same evaluation metrics as COCO-Position, with evaluations performed by the number of instances.

On the DrawBench dataset, we measure RaDL performance across three aspects: attribute assessment based on HSV color coverage, quantity assessment by comparing the number of generated instances to the prompt, and relation assessment by extracting instance centroids and comparing spatial relations in the image with those described in the text.

\begin{table}[tb]
    \centering
    \caption{Quantitative performance on the drawbench dataset}
    \label{tab:3}
    \resizebox{\columnwidth}{!}{
    \begin{tabular}{c|c|c|c}
    \toprule
    \textbf{Method} & \textbf{Attribute(\%) $\uparrow$} & \textbf{Quantity(\%) $\uparrow$} & \textbf{Relation(\%) $\uparrow$} \\
    \midrule
    GLIGEN & 51.00 & 44.08 & 60.62 \\
    BoxDiff & 28.50 & 9.21 & 31.25 \\
    TFLCG & 35.00 & 15.79 & 35.42 \\
    MultiDiffusion & 18.5 & 17.76 & 21.88 \\
    MIGC & 79.00 & 67.76 & 60.83 \\
    \midrule
    \rowcolor{Gray!25}
    Ours & \textbf{80.83} & \textbf{68.12} & \textbf{73.54} \\
    \bottomrule
    \end{tabular}
    }
\end{table}

\subsection{Quantitative Evaluation}

On the COCO-Position dataset, RaDL demonstrates the effectiveness of positional information in guiding image generation. It achieves an improved FID score of 23.53, surpassing the baseline score of 24.52, which is comparable to Stable Diffusion and indicative of high image quality. Furthermore, the instance success rate increases from 80.29\% to 81.09\%, exhibiting an overall enhancement in the accuracy of the multi-instance generation.

For the COCO-MIG benchmark, RaDL consistently outperforms the baselines in both Success Rate ($\text{R}$) for each instance and mIoU. In this setting, $L_i$ denotes the number of instances to be generated in the image. As the number of required instances increases, RaDL shows improvements from 56.88/49.89 to 59.45/53.27, demonstrating robust control over position, quantity, and multiple attributes.

In the DrawBench evaluation, RaDL exhibits performance gains across diverse conditions such as color, quantity, and relationships. Especially notable is the improvement in relationship understanding, where the accuracy rises significantly from 60.83\% to 73.54\%, highlighting the ability of the model to effectively consider relationships between instances.

\begin{figure}
    \centering
    \includegraphics[width=0.9\linewidth]{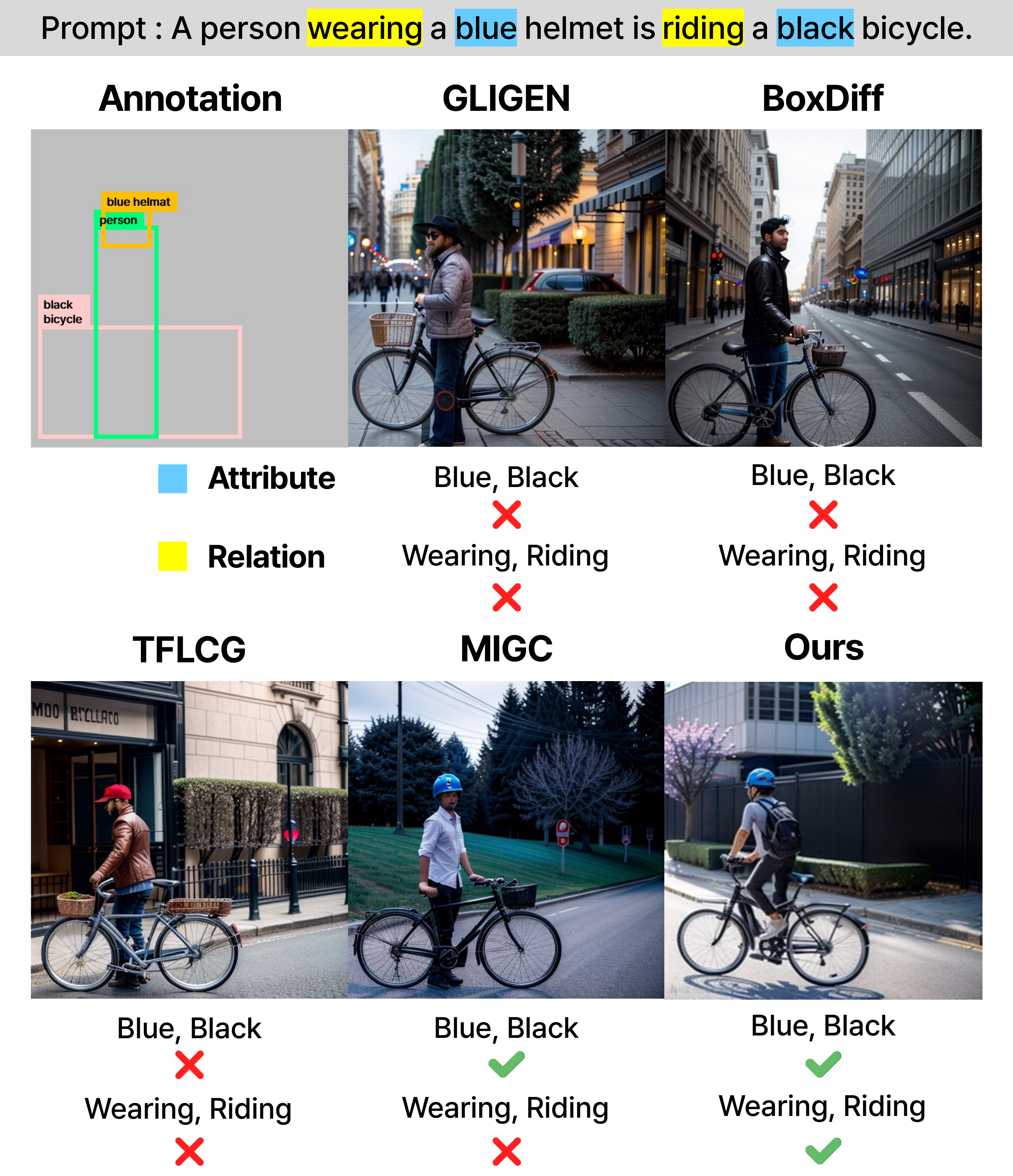}
    \caption{Qualitative comparison of DrawBench dataset. RaDL generates images, unlike existing models, that align with the context of the prompt while accurately considering the attributes and relationships between instances. The results can be seen in the generated images, where the person is correctly wearing a blue helmet and riding a black bicycle, with the attributes and the relations between the instances being matched.}
    \label{fig:4}
\end{figure}

\subsection{Qualitative Evaluation}

In Fig. \ref{fig:3}, we show that RaDL effectively generates instances within the designated boxes, preserving all specified attributes. While other models fail to account for multiple attributes within the color-coded boxes, RaDL demonstrates its ability to accurately consider and maintain these multiple attributes, ensuring that each instance considers its intended properties.

As shown in Fig. \ref{fig:4}, RaDL also exhibits precise generation over individual instance attributes (\textit{wearing a blue helmet} or \textit{riding a black bicycle}) as well as the relationships between instances. RaDL demonstrates that it effectively considers the semantics of the given prompts and is superior in generating images with multiple attributes and relationships.

\section{CONCLUSION}

In this paper, we propose RaDL to improve multi-instance image generation by focusing on both preservation of multiple attributes for each instance and consideration of relationships between instances. RaDL maintains and strengthens multiple attributes of each instance during training through Attribute Enhancement, ensuring that unique visual information is preserved. In addition, Relation Attention effectively addresses relationship discrepancy by extracting action verbs from the global prompt, allowing for proper representation of the relationships between each instance. The superior performance of RaDL across various benchmarks, including COCO-Position, COCO-MIG, and DrawBench, demonstrates its effectiveness in addressing the complex multi-instance image generation task while considering both attribute accuracy and relationships. We focus on generating images using bounding boxes, but future work will aim to apply additional layout structures, such as segmentation masks or keypoints, to improve spatial accuracy.

\addtolength{\textheight}{-12cm}   








\bibliographystyle{IEEEtran}
\bibliography{REFERENCE}

\end{document}